\documentclass[11pt]{article}

\usepackage[hyperref]{ACL2023}

\usepackage{color}

\usepackage{times}
\usepackage{latexsym}

\usepackage{bm}
\usepackage{array}
\usepackage{enumitem}
\usepackage{amsmath}
\usepackage{amsfonts}
\usepackage{multirow}
\usepackage{graphicx}
\usepackage{microtype}
\usepackage{subfigure}
\usepackage{appendix}
\usepackage{url}

\usepackage[T1]{fontenc}
\usepackage[utf8]{inputenc}

\usepackage{microtype}

\usepackage{inconsolata}

\title{MEMD-ABSA: A Multi-Element Multi-Domain Dataset for Aspect-Based Sentiment Analysis}
\author{
Hongjie Cai,
\hspace{2pt} Nan Song,
\hspace{2pt} Zengzhi Wang,
\hspace{2pt} Qiming Xie,
\hspace{2pt} Qiankun Zhao,
\hspace{2pt} Ke Li, \\
\hspace{2pt} {\bf Siwei Wu },
\hspace{2pt} {\bf Shijie Liu},
\hspace{2pt} {\bf Jianfei Yu},
\hspace{2pt} {\bf Rui Xia}\thanks{\hspace{0.3em} Corresponding author.}\\
School of Computer Science and Engineering,\\
Nanjing University of Science and Technology, China\\
\texttt{\{hjcai, nsong, zzwang, qmxie, kkzhao, kli, } \\
\texttt{wusiwei, sjliu, jfyu, rxia\}@njust.edu.cn}\\
}

\date{}

\begin{document}
\maketitle

\begin{abstract}

Aspect-based sentiment analysis is a long-standing research interest in the field of opinion mining, and in recent years, researchers have gradually shifted their focus from simple ABSA subtasks to end-to-end multi-element ABSA tasks. However, the datasets currently used in the research are limited to individual elements of specific tasks, usually focusing on in-domain settings, ignoring implicit aspects and opinions, and with a small data scale. To address these issues, we propose a large-scale Multi-Element Multi-Domain dataset (MEMD) that covers the four elements across five domains, including nearly 20,000 review sentences and 30,000 quadruples annotated with explicit and implicit aspects and opinions for ABSA research. Meanwhile, we evaluate generative and non-generative baselines on multiple ABSA subtasks under the open domain setting, and the results show that open domain ABSA as well as mining implicit aspects and opinions remain ongoing challenges to be addressed. The datasets are publicly released at \url{https://github.com/NUSTM/MEMD-ABSA}.

\end{abstract}

\section{Introduction}

\begin{figure}[!tp]
	\centering
	\begin{tabular}{c}
		\hspace{-0.25cm}\hbox{\includegraphics[scale=0.2]{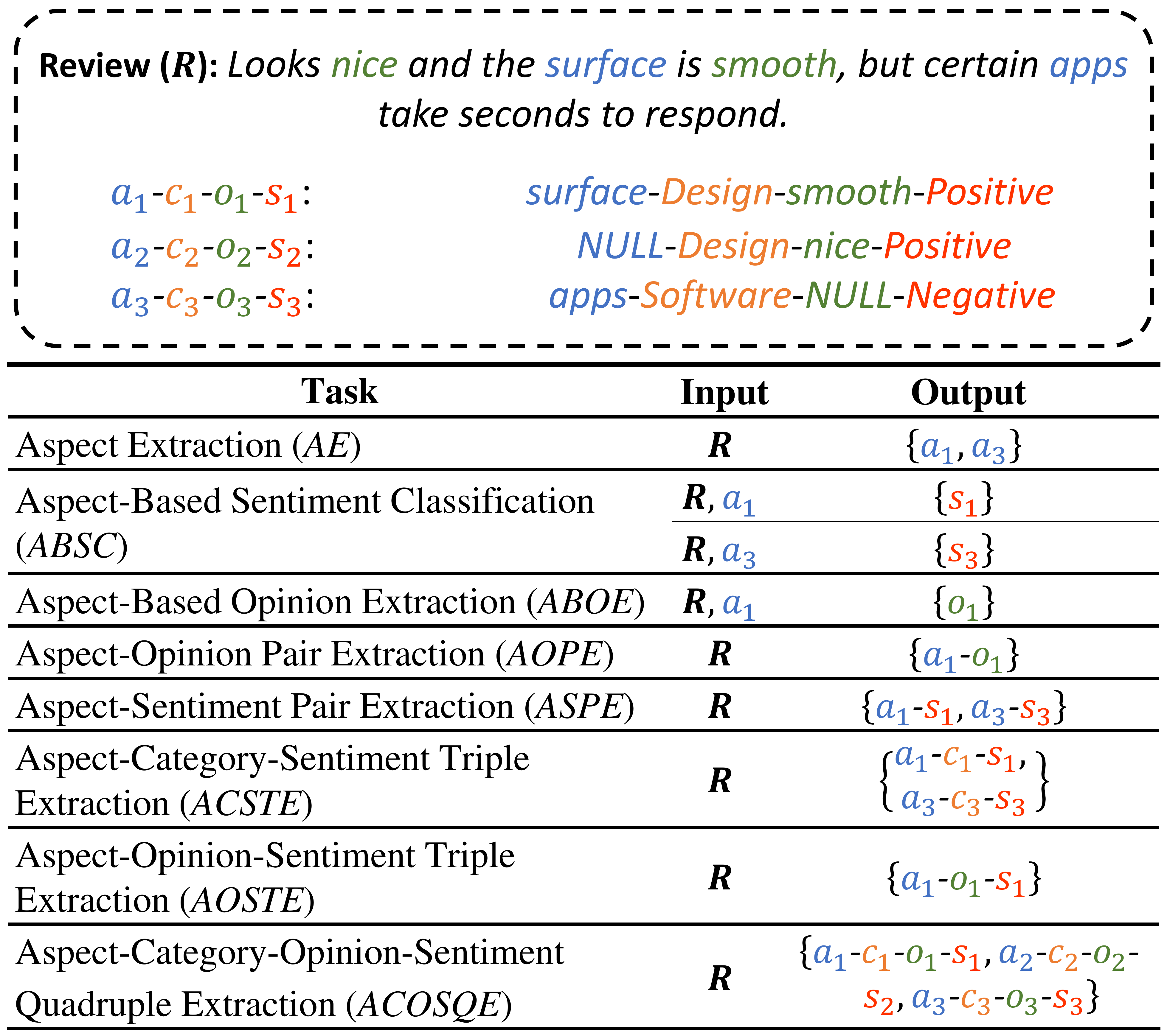}}
	\end{tabular}
	\caption{Illustration of the input and output of ABSA tasks.}
	\label{fig:absa_task_input_output}
\end{figure}

%As a fine-grained sentiment analysis task, Aspect-Based Sentiment Analysis (ABSA) aims to identify the opinion target and the expressed opinion towards the target from text reviews \cite{liu2012sentiment}. 
Aspect-Based Sentiment Analysis (ABSA) is an important task for fine-grained sentiment analysis, which aims to extract the
opinion target described by an entity and its aspect
(collectively called aspect) from reviews and identify the sentiment toward the aspect~\cite{liu2012sentiment}.
The ABSA task typically consists of four core elements: 
%\textit{Aspect}, \textit{Category}, \textit{Opinion}, and \textit{Sentiment}. Among them, \textit{Aspect} represents an entity or a specific aspect of the entity with opinion expressions. \textit{Category} denotes a predefined category describing the entity or its aspect. \textit{Opinion} represents the opinion expression regarding the \textit{Aspect}, while \textit{Sentiment} indicates the sentiment polarity towards the \textit{Aspect}. For example, as mentioned in Figure~\ref{fig:quadruple_example}, "the surface is smooth," \textit{Aspect} refers to "surface", its corresponding \textit{Category} is Design, the \textit{Opinion} towards "surface" is smooth, and its \textit{Sentiment} is Positive.
1) \textit{{\color{blue}Aspect term}} denoting the entity or aspect terms on which opinions have been expressed; 
2) \textit{\color{orange}Aspect category} representing the pre-defined category of concerned aspects;
3) \textit{\color{green}Opinion term} denoting the opinion words or phrases towards aspects;
4) \textit{\color{red}Sentiment polarity} representing the sentiment polarities towards aspects.
For example, the four elements of the sentence fragment ``\textit{the surface is smooth}'' in figure~\ref{fig:absa_task_input_output} are \textit{\color{blue}surface}, \textit{\color{orange}Design}, \textit{\color{green}smooth}, and \textit{\color{red}Positive}, respectively.

% 作为一个细粒度观点挖掘任务，属性级情感分析（ABSA）旨在从评论文本中识别评价对象以及针对评价对象表达的观点信息\cite{liu2012sentiment}，该任务通常包含四个核心要素：Aspect， Category， Opinion， Sentiment。其中，我们用Aspect表示有相关情感表达的entity或entity的某个aspect；Category表示预定义好的描述entity及其aspect的标签集合；Opinion表示针对Aspect的情感表达，Sentiment表示表达在Aspect上的预定义好的情感极性。如figure~\ref{fig:quadruple_example}中提到的：“the surface is smooth”，Aspect是“sushi”，对应的Category是Design，相关的情感表达是smooth，它的情感极性是Positive。

During the past decade, the ABSA task has attracted wide attention from both academia and industry, ranging from the traditional single-element extraction 
to the recent multi-element extraction.
%ABSA typically includes single-element extraction and multi-element extraction tasks. Along with these tasks, there are many benchmark datasets containing fine-grained opinion annotation, such as \cite{hu2004mining, mitchell2013open, zhao2014creating, pontiki-etal-2014-semeval, pontiki2015semeval, pontiki2016semeval}.
% ABSA typically包含单要素抽取/识别和多要素抽取/识别任务。伴随着这些任务出现的，还有许多包含细粒度评价对象情感标注的benchmark数据集\cite{hu2004mining,mitchell2013open,zhao2014creating,pontiki-etal-2014-semeval,pontiki2015semeval,pontiki2016semeval}，
%It is worth noting that these 
The early benchmark datasets were primarily used to assist research on single-element extraction tasks. In recent years, as the focus of ABSA research has gradually shifted from single-element extraction to multi-element extraction, more datasets for multi-element extraction have been proposed, including \cite{wang2016recursive, wang2017coupled, fan2019target, xu2020position}. Figure~\ref{fig:absa_task_input_output} illustrates eight representative ABSA tasks, including AE, ABSC, ABOE, AOPE, ASPE, ACSTE, AOSTE, ACOSQE, etc\footnote{Some tasks may be referred to as other appellations in prior works, e.g., ABOE is also known as Target-oriented Opinion Words Extraction (TOWE). For the consistency in terminology, we refer to it as ABOE in this work. The same goes for other tasks.}.

%Considering that close to 40\% of sentences contain implicit aspects or opinions, \cite{cai2021aspect} proposed a four-element dataset that includes explicit and implicit aspects and opinions. Meanwhile, many fully supervised approaches have been proposed to address multi-element extraction tasks, such as double-element extraction \cite{li2019unified, chen2020synchronous, schmitt2018joint}, triple-element extraction \cite{peng2020knowing, wan2020target}, and complete four-element extraction \cite{cai2021aspect, zhang2021aspect}. Recently, with the emergence of Large Language Models represented by ChatGPT, there has been work on evaluating single-element and multi-element tasks using in-context learning (ICL) \cite{brown2020language, min2022rethinking} \cite{wang2023chatgpt}.
% 值得注意的是，早期的这些benchmark数据集主要用来辅助针对单要素抽取任务的研究。近年来，随着ABSA领域的研究重心逐渐从单要素抽取转移到多要素抽取，更多多要素抽取数据集\cite{wang2016recursive, wang2017coupled, fan2019target,xu2020position}被推出；考虑到接近40\%的句子中包含隐式Aspect或Opinion，\cite{cai2021aspect}提出了包含显式和隐式标注的四要素数据集。同时，许多基于全监督的方法被提出以解决多要素抽取任务，如，双要素抽取\cite{li2019unified,chen2020synchronous,schmitt2018joint}，三要素抽取\cite{peng2020knowing,wan2020target}，以及完整的四要素抽取\cite{cai2021aspect,zhang2021aspect}。

Despite the extensive research on the aforementioned ABSA tasks, the following issues persist in this field:
% 尽管上述这些ABSA任务和方法已经被广泛讨论，然而这个领域仍存在以下问题：
%$($1$)$ 从数据角度来看，首先，现有的针对多要素任务的研究主要集中在Restaurant和Laptop领域训练评估，涵盖的领域少，标注的样例规模小，并且，现有标注数据的标签很难完整刻画复杂且多样的细粒度评价对象及其情感表达，尤其是在多要素任务中；其次，大多数ABSA数据集只考虑了显式评价对象和情感表达，忽略了隐式的。尽管\cite{cai2021aspect}提出了Aspect-Category-Opinion-Sentiment （ACOS）四要素抽取任务，并提供了Restaurant和Laptop领域的隐式评价对象和情感表达标注，如figure~\ref{fig:quadruple_example}所示，给定评论文本，ACOS任务不仅要抽取出显式的四元组(surface-Design-smooth-Positive)，还要抽取出包含隐式评价对象的四元组(NULL-Design-nice-Positive)以及隐式情感表达的四元组(apps-Software-NULL-Negative)。可以看到，隐式评价对象和情感表达的文本表达比显式的更加丰富，他们在不同领域的差异会更大，并且已有的隐式标注规模仍然较小。
Firstly, existing studies were mainly conducted and evaluated on two small datasets with limited annotations from the \textit{Restaurant} and \textit{Laptop} domains, which were introduced by the SemEval 2014-2016 Shared Tasks ~\cite{pontiki-etal-2014-semeval, pontiki2015semeval, pontiki2016semeval}.
In recent years, the datasets for multi-element tasks have also been supplemented and annotated based on the existing datasets, and the scale is relatively small.
% 近几年多要素任务的数据集也大多在这两个领域的数据上进行了补充标注，并且规模都比较小。
However, training deep learning models with millions of parameters, on such small-scale datasets, tends to bear the risk of over-fitting. It is hence inadequate to clearly compare different models and reflect their performance in the open domain. 
It is worth noting that Large Language Models (LLMs), represented by ChatGPT, have had an impact on aspect-based sentiment analysis tasks \cite{wang2023chatgpt}. Although they do not rely on domain-specific annotated data for training, evaluation of LLMs' open-domain sentiment analysis capabilities across multiple domains is still needed.
% it is still necessary to evaluate the open-domain sentiment analysis capabilities of LLMs across multiple domains.
% 值得注意的是，以Chatgpt为代表的LLMs给属性级情感分析任务带来了冲击\cite{wang2023chatgpt}。尽管它们不依赖于特定领域标注数据进行模型的训练，但是仍需要在多个领域上去评估LLMs的开放域情感分析能力。
%的出现，开始有基于 in-context
% learning (ICL) \cite{brown2020language,min2022rethinking}来评估单要素和多要素任务的工作
Therefore, there still lacks an impactful large-scale multi-domain multi-element dataset for the ABSA tasks.

Secondly, most existing ABSA datasets only considered the extraction of explicit aspects and opinions, while ignoring the implicit ones.
According to the statistics in \cite{zhao2014creating,cai2021aspect},
%\citet{zhao2015creating} and \citet{cai2021aspect}, 
around 40\% of the review sentences contain implicit aspects or implicit opinions. 
For example in figure~\ref{fig:absa_task_input_output}, “looks nice” does not contain an explicit aspect term, yet it points to an implicit aspect regarding the Design of the phone. Similarly, a sentence that does not contain an explicit opinion term (e.g., “certain apps take seconds to respond”) often expresses a negative or positive sentiment. Such implicit aspect or opinion occur frequently in product reviews and are very important for ABSA, 
%at a very high rate 
however, were normally ignored in both research and applications. 
To tackle this problem, our pilot study~\cite{cai2021aspect} proposed an Aspect-Category-Opinion-Sentiment (ACOS) quadruple extraction task, aiming to extract all aspect-category-opinion-sentiment quadruples with both explicit and implicit aspects and opinions in a review. 
Given the review in Fig.~\ref{fig:absa_task_input_output}, the task requires extracting all three ACOS quadruples, including one explicit quadruple (\textit{\color{blue}surface}-\textit{\color{orange}Design}-\textit{\color{green}smooth}-\textit{\color{red}Positive}), one with an implicit aspect (\textit{\color{blue}NULL}-\textit{\color{orange}Design}-\textit{\color{green}nice}-\textit{\color{red}Positive}) and one with an implicit opinion (\textit{\color{blue}apps}-\textit{\color{orange}Software}-\textit{\color{green}NULL}-\textit{\color{red}Negative}).

%$($2$)$ 从模型角度来看，在少领域小规模的数据集上训练参数量百万级的模型不仅容易出现过拟合的问题，也限制了模型的泛化能力。
%更大参数规模的语言模型\cite{brown2020language,hoffmann2022training,ouyang2022training}虽然在few-shot setting下表现更好，但相比全监督训练的结果仍然相差很多,如Table~\ref{tab:ACOS comparison}所示，我们在ACOS任务的测试集中选取了30条包含显隐式Aspect和Opinion的测试样例，Chatgpt在5-shot setting下的结果比全监督的结果低20\%左右。
%综上所述，目前ABSA仍是仍然充满挑战，且该领域仍然缺少覆盖多个领域，包含多种要素和隐式要素标注的数据集。
%\cite{chia2023domain}提供了三元组的数据集。

% \begin{table}[!tp]
% 	\small
% 	%	\renewcommand\tabcolsep{7.0pt} % ?????????
	
% 	\centering
% 	\resizebox{\linewidth}{!}{
% 	\setlength{\tabcolsep}{2mm}{
% 		\begin{tabular} {l|c|c}
% 			\hline
% 			& 5-shot & \small Fully supervised \\
% 			\hline \hline
% 			Chatgpt & 44.19\% & - \\
% 			\hline
% 			T5-Paraphrase & 7.41\% & 65.30\% \\
%                 \hline
% 	\end{tabular}}}
% 	\caption{Comparison results of Chatgpt and T5-Paraphrase\cite{zhang2021aspect} on 5-shot and fully supervised setting.}
% 	\label{tab:ACOS comparison}
% \end{table}

To address these limitations, we propose a new ACOS dataset that covers five domains: Book, Clothing, Hotel, Restaurant, and Laptop. The dataset contains nearly 20,000 review sentences, which is four to five times larger than the previous SemEval ABSA datasets. In addition, it is annotated with almost 30,000 quadruples, supporting multi-element extraction tasks involving both explicit and implicit aspects and opinions.
Furthermore, considering that most previous studies were conducted under in-domain settings, we comprehensively evaluate various typical generative and non-generative methods in an open-domain setting and analyze their performance on different subtasks of the new dataset.

% 为了解决这些问题，我们提出了一个新的四元组数据集，该数据集涵盖Book，Clothing，Hotel，Restaurant以及Laptop五个领域，共包含近20000条评论句，比先前的SemEval ABSA数据集大4到5倍。同时，该数据集标注了接近30000个的四元组，支持包含显隐式aspect和opinion的多要素抽取任务。
% 更进一步，考虑到先前研究大多在in-domain设定下开展，我们在open-domain设定下，对多种典型的生成式和非生成式方法进行了全面评估，并分析了他们在新数据集的各个子任务上的表现。
% 通过这一研究，我们希望借助这个四元组数据集推动相关领域的技术研究和创新，进一步提升模型在处理实际应用场景时的准确性和可靠性。同时，希望能深入探讨各类方法在不同领域下的适用性和性能表现，从而为用户提供更加系统化、全面的参考依据。

\begin{table*}[!tp]
% 	\vspace{-0.1em}
	\footnotesize
    \centering
    % \small
	%\setlength{\tabcolsep}{0.5em}
	\setlength{\abovecaptionskip}{0.2cm}
  	\setlength{\belowcaptionskip}{-0.1cm}
	\setlength{\tabcolsep}{0.8mm}
	\resizebox{\textwidth}{!}
	{\renewcommand{\arraystretch}{1.2}
		\begin{tabular}{ll|l|c|c|c|c|c|c|c|c|c|c}
            % \multirow{5}{*}{}
			\hline
			 & & & \multirow{2}{*}{\textbf{Sentence}} & \multirow{2}{*}{\textbf{Aspect}} & \multirow{2}{*}{\textbf{Category}} & \multirow{2}{*}{\textbf{Opinion}} & \multirow{2}{*}{\textbf{Sentiment}} & \textbf{AS} & \textbf{AO} & \textbf{AOS} & \textbf{ACS} & \textbf{ACOS} \\
			
		&&	& & & & & & \textbf{Pair} & \textbf{Pair} & \textbf{Triple} & \textbf{Triple} & \textbf{Quadruple} \\
			\hline\hline
			
	\multirow{4}{*}{\rotatebox{90}{\textbf{Previous}}}& \multirow{4}{*}{\rotatebox{90}{\textbf{SemEval}}}	&	Restaurant-2014  & \makebox[5ex][r]{3841} & \makebox[5ex][r]{4827} & \makebox[5ex][r]{4738} & \text{-} & \makebox[5ex][r]{4534} & \makebox[5ex][r]{4827} & \text{-} & \text{-} & \text{-} & \text{-} \\
			\cline{3-13}
			
		&	& Laptop-2014  & \makebox[5ex][r]{1910} & \makebox[5ex][r]{3012} & \text{-} & \text{-} & \makebox[5ex][r]{3012} & \makebox[5ex][r]{3012} & \text{-} & \text{-} & \text{-} & \text{-} \\
			\cline{3-13}
			
		&	& Restaurant-2016  & \makebox[5ex][r]{2295} & \makebox[5ex][r]{3122} & \makebox[5ex][r]{3001} & \text{-} & \makebox[5ex][r]{3122} & \makebox[5ex][r]{3182} & \text{-} & \text{-} & \makebox[5ex][r]{3364} & \text{-} \\
			\cline{3-13}
			
	    &	& Laptop-2016 & \makebox[5ex][r]{2612} & \text{-} & \makebox[5ex][r]{3705} & \text{-} & \makebox[5ex][r]{3705} & \text{-} & \text{-} & \text{-} & \text{-} & \text{-} \\
			\hline\hline
	\multirow{6}{*}{\rotatebox{90}{\textbf{Our}}}& \multirow{6}{*}{\rotatebox{90}{\textbf{Annotation}}}& Books & \makebox[5ex][r]{2967} & \makebox[5ex][r]{3781} & \makebox[5ex][r]{3593} & \makebox[5ex][r]{4291} & \makebox[5ex][r]{3781} & \makebox[5ex][r]{3931} & \makebox[5ex][r]{4493} & \makebox[5ex][r]{4493} & \makebox[5ex][r]{4048} & \makebox[5ex][r]{4507} \\
			\cline{3-13}
			
		&	& Clothing & \makebox[5ex][r]{2373} & \makebox[5ex][r]{2843} & \makebox[5ex][r]{2994} & \makebox[5ex][r]{3341} & \makebox[5ex][r]{2843} & \makebox[5ex][r]{2904} & \makebox[5ex][r]{3415} & \makebox[5ex][r]{3415} & \makebox[5ex][r]{3186} & \makebox[5ex][r]{3416} \\
			\cline{3-13}
			
		&	& Hotel & \makebox[5ex][r]{3526} & \makebox[5ex][r]{4700} & \makebox[5ex][r]{4886} & \makebox[5ex][r]{5781} & \makebox[5ex][r]{4700} & \makebox[5ex][r]{4735} & \makebox[5ex][r]{6014} & \makebox[5ex][r]{6014} & \makebox[5ex][r]{5284} & \makebox[5ex][r]{6017} \\
			\cline{3-13}
			
		&	& Restaurant & \makebox[5ex][r]{5152} & \makebox[5ex][r]{7056} & \makebox[5ex][r]{6307} & \makebox[5ex][r]{7958} & \makebox[5ex][r]{7056}
            & \makebox[5ex][r]{7250} & \makebox[5ex][r]{8484} & \makebox[5ex][r]{8484} & \makebox[5ex][r]{7436} & \makebox[5ex][r]{8496} \\
            \cline{3-13}
        &    & Laptop & \makebox[5ex][r]{4076} & \makebox[5ex][r]{4958} & \makebox[5ex][r]{4992} & \makebox[5ex][r]{5378} & \makebox[5ex][r]{4958} & \makebox[5ex][r]{5035} & \makebox[5ex][r]{5726} & \makebox[5ex][r]{5731} & \makebox[5ex][r]{5227} & \makebox[5ex][r]{5758} \\
			\cline{3-13}
		&	& Total & \makebox[5ex][r]{18094} & \makebox[5ex][r]{23338} & \makebox[5ex][r]{22772} & \makebox[5ex][r]{26749} & \makebox[5ex][r]{23338} & \,\makebox[5ex][r]{23855} & \,\makebox[5ex][r]{28132} & \,\makebox[5ex][r]{28137} & \,\makebox[5ex][r]{25181} & \makebox[5ex][r]{28194} \\
			\hline
	\end{tabular}}
	
	\caption{Statistics of our annotation.
	AS, AO, AOS, ACS, and ACOS denote aspect-sentiment, aspect-opinion, aspect-opinion-sentiment, aspect-category-sentiment, and aspect-category-opinion-sentiment, respectively.}
	\label{tab:dataset_statistic}
\end{table*}

\section{Multi-Element Multi-Domain ABSA Datasets}
% (with category and sentiment Explanations)

\subsection{Data Collection}

To construct the ACOS dataset, we collect raw data from five different social media domains and annotate quadruples of these raw data. The data sources are as follows: The Books and Clothing domains data are obtained from the 5-core version provided by \citet{ni2019justifying}\footnote{\url{https://nijianmo.github.io/amazon/}}.
We randomly select 986 book reviews and 928 clothing reviews from them.
The Restaurant and Hotel domain data are collected from the Boston dataset of Yelp\footnote{\url{https://www.yelp.com/dataset/download}}
and the Boston dataset of Airbnb\footnote{\url{http://insideairbnb.com/get-the-data/}}, respectively. We randomly select and annotate 940 and 1029 reviews from the two datasets, respectively. The Laptop domain dataset is selected from the Amazon platform\footnote{\url{https://www.amazon.com/}}
and includes product reviews from 2017 and 2018 covering six laptop brands (ASUS, Acer, Samsung, Lenovo, MBP, MSI). We use nltk to tokenize the selected reviews at the sentence and word levels and use the processed sentences for subsequent annotation.

\subsection{Data Annotation}

In this section, we will introduce the annotation guidelines and process, respectively. Considering the need to label text spans and their relations, we use the Inception platform \cite{klie2018inception} to annotate the ACOS quadruples.

The Inception platform can be used to annotate various natural language processing tasks in text form.
Before annotation, we import the raw corpus for each domain, add the tagsets for the categories and sentiments corresponding to each domain, and define the relation type between aspect and opinion. 
\begin{figure*}[!tp]
	\centering
	\begin{tabular}{c}
		\hspace{-0.25cm}\hbox{\includegraphics[scale=0.4]{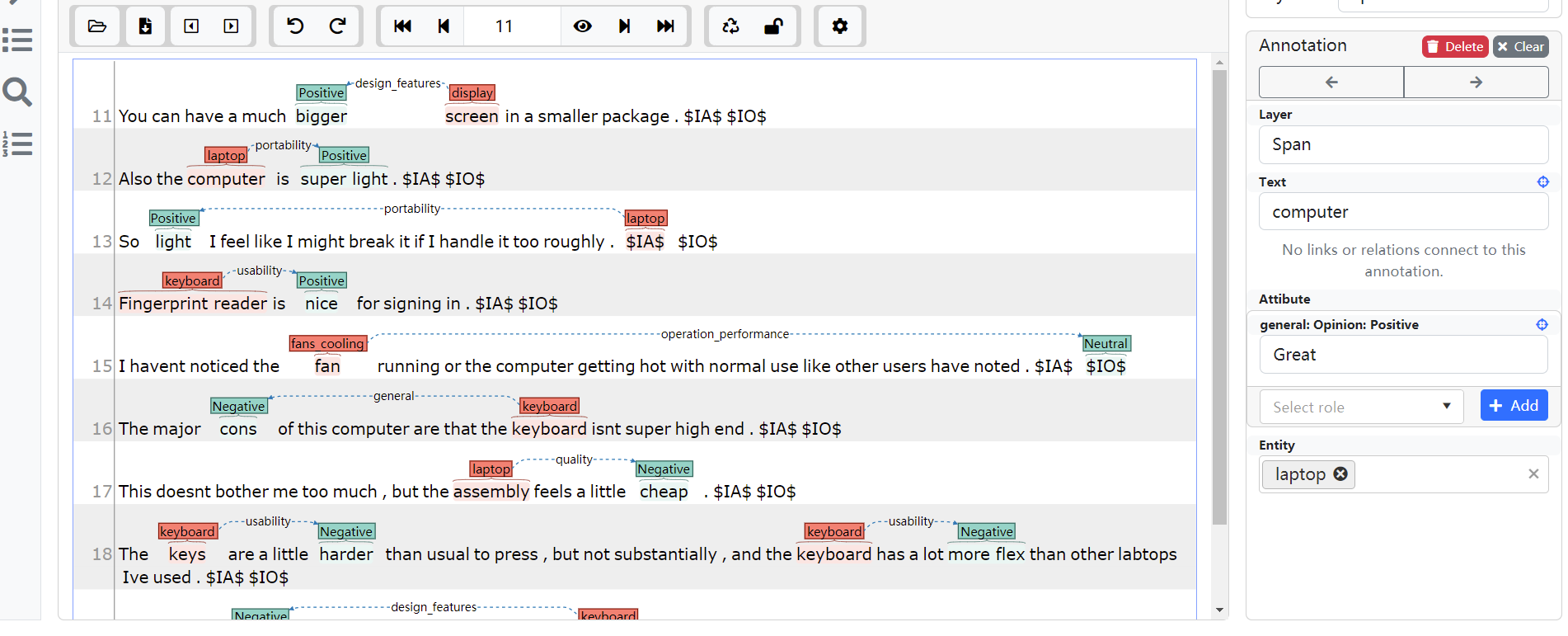}}
	\end{tabular}
	\caption{The annotation interface of Inception for the ACOS quadruple extraction task.}
	\label{fig:anntation_interface}
\end{figure*}

Figure~\ref{fig:anntation_interface} shows the annotation interface and some annotated examples of the laptop domain dataset using Inception for annotation.
% We will show the interface for annotating quadruples using Inception in the appendix \ref{appendix:annotation}.

\subsubsection{Annotation Guidelines}

We formalize the ACOS extraction task as a structured information extraction task. In order to annotate both explicit and implicit aspects and opinions, and pair them with category and sentiment to form complete quadruples, we provide the specific definition of the four elements in the review sentence:

\begin{itemize}[leftmargin=0.15in]
\setlength\itemsep{-0.2em}
\item \textit{{\color{blue}Aspect term}} abbreviated as Aspect, refers to the word or phrase in the sentence, including explicit and implicit aspects. Specifically, we add a special token "IA" after each sentence to indicate implicit aspects.
\item \textit{\color{orange}Aspect category} abbreviated as Category, selected from a predefined set of category tags, where each category tag consists of an entity tag (such as Laptop) and an attribute tag (such as Design).
\item \textit{\color{green}Opinion term} abbreviated as Opinion, refers to the word or phrase in the sentence, including explicit and implicit opinion. Specifically, we add a special token "IO" after each sentence to indicate implicit opinion.
\item \textit{\color{red}Sentiment polarity} abbreviated as Sentiment, selected from predefined sentiment polarity set: \{Positive, Negative, Neutral\}.
\end{itemize}

\paragraph{Aspect Annotation}

We set the following principles for aspect annotation: 1) The aspect span should correspond to a complete objective thing; 2) The span should cover the longest effective description of the aspect. For example, "Huawei's Mate 30 Phone" or other proper nouns correspond to both an objective thing and the longest effective aspect description. Articles (such as "the", "a", "an", "their", etc.) are not included in the aspect. Specific examples of aspect annotation are shown in the appendix \ref{appendix:annotation}.

\paragraph{Category Definition}

Categories of different domains have different sources. Categories of the Hotel, Restaurant, and Laptop domains are from \cite{pontiki2016semeval}. Categories of the Books domain are selected from \cite{alvarez2018proposal}. To determine the categories of the Clothing domain, we ask two annotators to read 100 Clothing domain review texts and summarize the entity tags and attribute tags that appeared. The two annotators then discuss with a third expert to determine the final category tags. Detailed category tags for each domain can be found in the appendix \ref{appendix:category_definition}.

\paragraph{Opinion Annotation}

We set the following principles for opinion annotation: annotate the opinion words (usually adjectives, adverbs, verbs, and some nouns) or phrases that indicate the sentiment polarities of the reviewer of the sentence. For difficult-to-determine opinions, we refer to the MPQA Subjectivity Lexicon \cite{wilson2005recognizing} to determine whether to label the span as an opinion, otherwise, we consider whether there is an implicit opinion. In addition to opinion words, we also include polarity shifters (such as intensifiers, diminishers, and negations) \cite{liu2012sentiment,zong2021text} in opinion annotation.

\begin{figure}[!t]
	\centering
		\hspace{-0.2cm}\includegraphics[scale=0.36]{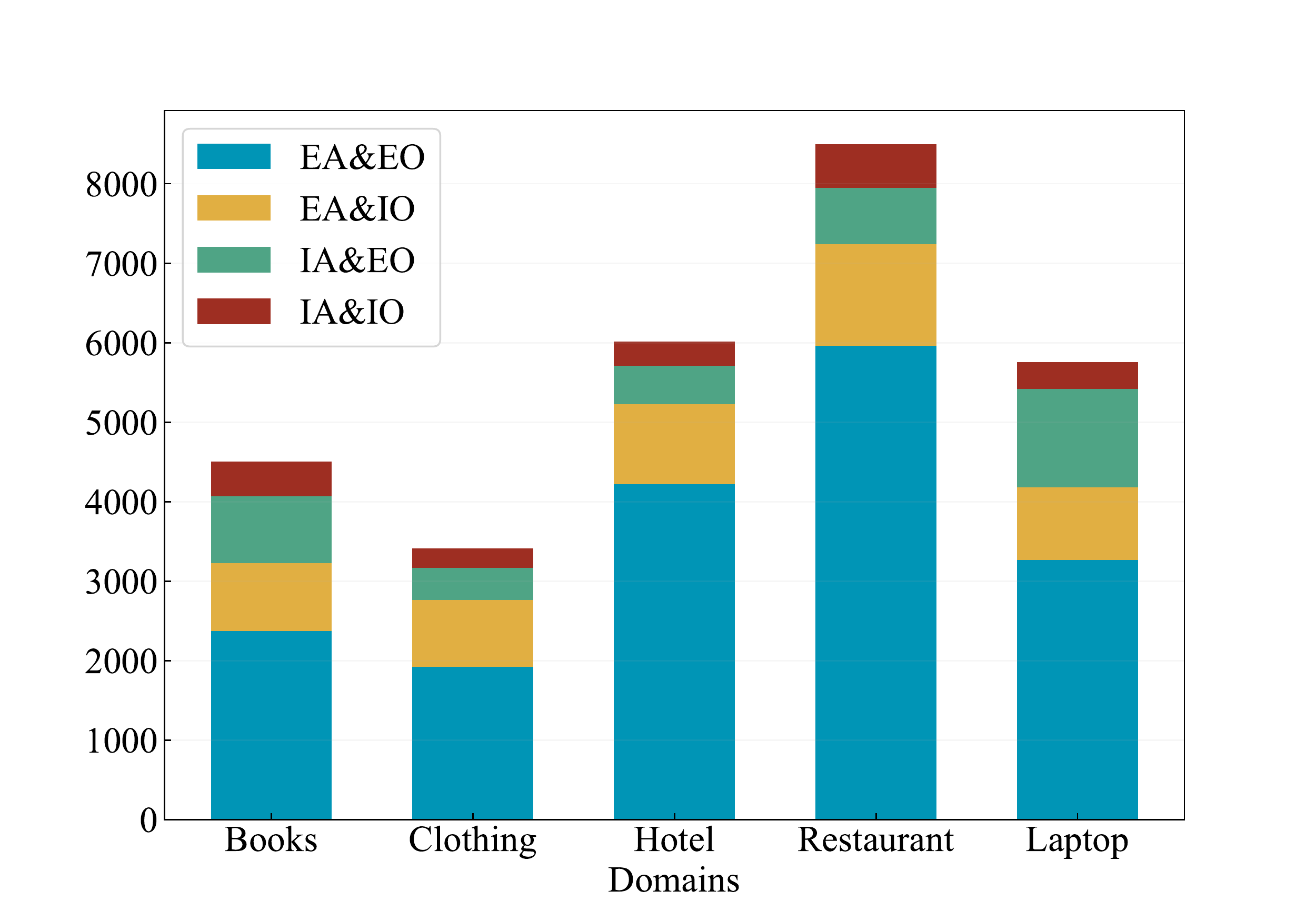}

	\caption{Number of explicit and implicit quadruples of five domains. EA, EO, IA, and IO denote explicit aspect, explicit opinion, implicit aspect, and implicit opinion, respectively.}
	\label{fig:exp_imp_percent}
\end{figure}

\begin{figure*}[!t]
	\centering
        \subfigure[Restaurant Domain] {
		\hspace{-0.48cm}
            \begin{minipage}[t]{\linewidth}
			\centering
        \includegraphics[scale=0.464]{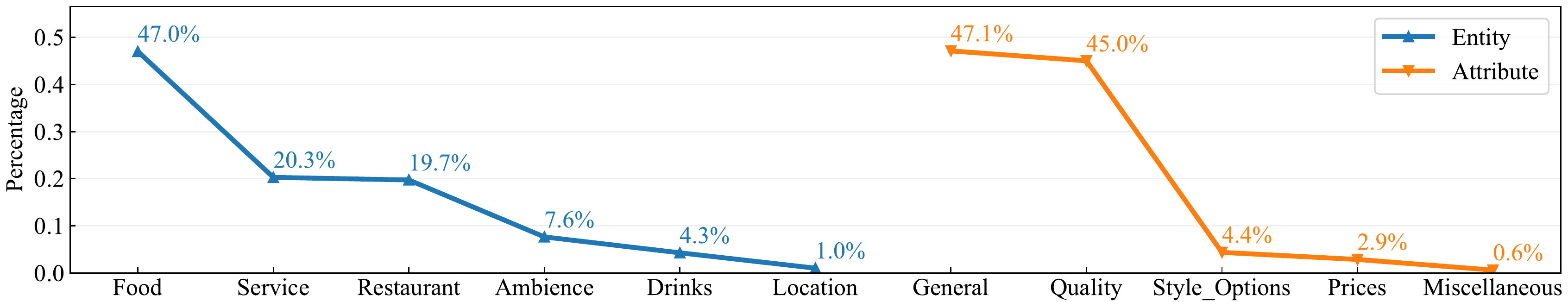}
			%\caption{fig2}
		\end{minipage}
	}
        % \vspace{0.1cm}
	   \subfigure[Clothing Domain] {
    	\hspace{-0.3cm}	
            \begin{minipage}[t]{\linewidth}
			\centering
			\includegraphics[scale=0.458]{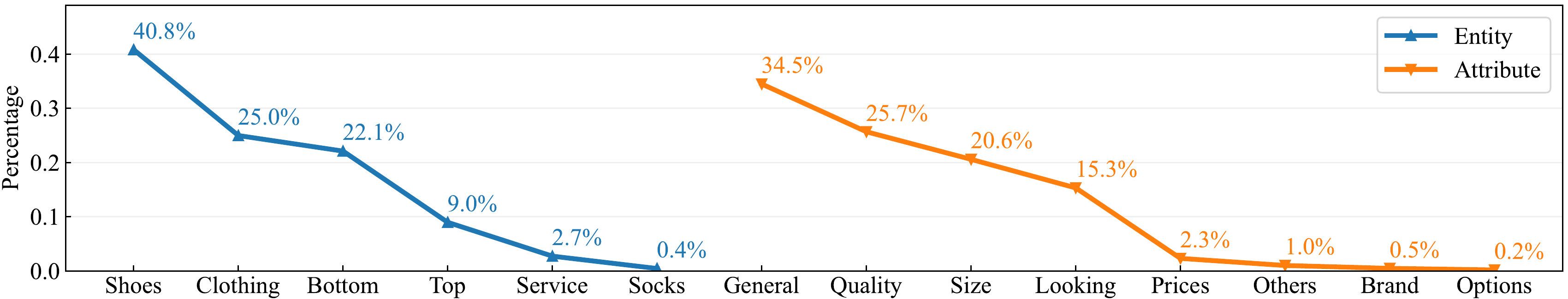}
			%\caption{fig2}
		\end{minipage}
        }
	\caption{Entity and attribute distribution of Restaurant and Clothing domains.}
	\label{fig:category_distribution}
\end{figure*}

\paragraph{Discontinuous Aspect/Opinion}

There are also some special expressions of aspect and opinion in the review sentences, which are not included in our annotation but are listed here. Discontinuous aspect refers to the text span of the aspect being discontinuous, in the text "Sauce has sesame seeds, and it is a perfect match with potato edge.", the aspect "sauce with potato edge," is discontinuous. Unlike the hierarchical opinion target defined by entities and aspects in \cite{liu2012sentiment}, in general, the internal entities of such discontinuous aspects do not have hierarchical relations. They are generally composed of continuous central words (or phrases, such as "sauce") and continuous auxiliary words (or phrases, such as "potato edge").

The discontinuous opinion refers to the text span of the opinion expression being discontinuous, usually in the form of polarity negation, in the text "I don't think the shirt is good.", the complete opinion is "don't good,", the central word of the opinion expression is "good," and the auxiliary word is "don't.". The difference between discontinuous opinion and implicit opinion is that implicit opinion does not have a central word of sentiment expression.

% 评论句中还会出现部分特殊表现形式的Aspect和Opinion，虽然我们未将它们纳入标注范畴，但也在此一并列出。不连续Aspect指有情感表达的Aspect对应的文本片段不连续，如文本"Sauce has sesame seeds, and it is perfect match with potato edge."，其中有情感表达的Aspect是sauce with potato edge，该评价对象是不连续的。不同于\cite{liu2012sentiment}定义的由entity和aspect构成的hierarchical评价对象，一般情况下，这种不连续评价对象的内部实体没有实际的层次关系。一般由连续的中心词(或短语，如：sauce)与连续的辅助词(或短语，如：potato edge)组成。
% 不连续的Opinion指情感表达对应的文本片段不连续，通常是否定形式的polarity shift，如文本"I don’t think the shirt is good."，完整的Opinion是"don't good",情感表达的中心词是"good"，辅助词是"don’t"，不连续Opinion与隐式Opinion一个明显的区别是，隐式Opinion没有情感表达的中心词。

\subsubsection{Annotation Process}

For each sentence, the annotators need to read the entire sentence and look for the aspect with opinion expression. For each of the aspects, first, the annotators need to mark the aspect as an Aspect and determine the entity tag of Category for the Aspect. Then, find the corresponding Opinion for the Aspect and pair them together. Since the attribute tag of the Category is generally determined by both the Aspect and the Opinion, especially when either the Aspect or the Opinion is implicit, the relation between the Aspect and the Opinion needs to be set as an attribute tag of the Category, forming a complete ACOS quadruple annotation. 
The average annotation F1 score for strict matching of quadruples in the domains is 68.52\%, indicating a high consistency in quadruple annotation across different domains. For annotation examples where two annotators cannot reach a consensus, a third expert is consulted to discuss and determine the final annotation.

\begin{table}[!tp]
	\scriptsize

	\centering
	\resizebox{\linewidth}{!}{
        \renewcommand{\arraystretch}{1.2}
	\setlength{\tabcolsep}{2mm}{
		\begin{tabular} {l|c|c|c}
			\hline
			& positive & negative & neutral \\
			\hline \hline
			Books & 70.47\% & 26.65\% & 2.88\% \\
			\hline
			Clothing & 74.74\% & 23.45\% & 1.81\% \\
			\hline
                Hotel & 95.53\% & 4.14\% & 0.33\% \\
			\hline
                Restaurant & 75.62\% & 23.01\% & 1.37\% \\
			\hline
                Laptop & 61.91\% & 32.62\% & 5.47\% \\
			\hline
	\end{tabular}}}
	\caption{Polarity distribution per domain.}
	\label{tab:polarity_distribution}
\end{table}

\subsection{Dataset Analysis}

We compare our dataset with the benchmark datasets provided by SemEval 2014 and 2016, as shown in Table~\ref{tab:dataset_statistic}, in addition to the Laptop and Restaurant domains annotated by SemEval, we also annotated data in three additional domains: Books, Clothing, and Hotel, which include all four elements and pairings. The size of the sentences is about three times that of the SemEval 2016 dataset, with nearly 30,000 quadruple annotations, and an average of 1.56 quadruples annotated per sentence. Furthermore, our annotated data can be used for not only the quadruple extraction task but also other subtasks of ABSA.

\paragraph{Implicit Aspect\&Opinion Statistics}

We count the number of quadruples in each domain for four types of aspect-opinion pair: explicit aspect-explicit opinion (EA\&EO), explicit aspect-implicit opinion (EA\&IO), implicit aspect-explicit opinion (IA\&EO), and implicit aspect-implicit opinion (IA\&IO), as shown in Figure~\ref{fig:exp_imp_percent}. It can be seen that the proportion of the four types of quadruples is almost consistent across all domains. Among them, EA\&EO accounts for about 60\% of all quadruples, followed by EA\&IO and IA\&EO, which together accounts for about 30\% of all quadruples, and IA\&IO has the lowest percentage, representing only about 7\% of all quadruples.

\paragraph{Category Distribution}

We count the proportions of entities and attributes in the Restaurant and Clothing domains. As shown in Figure~\ref{fig:category_distribution} (a), the most common entity in the Restaurant domain is Food, which accounts for nearly half of all entities, while Service and Restaurant are the next most discussed entities, each accounting for about 20\% of the total, and Ambience, Drinks, and Location account for about 10\% of the total. The most commonly discussed attributes in the Restaurant domain are General and Quality, which together account for approximately 90\%. 
The distribution of the entities in the Clothing domain is similar to that in the Restaurant domain. As shown in Figure~\ref{fig:category_distribution} (b), the most discussed entity is Shoes, which accounts for about 40\% of all entities, followed by Clothing and Bottom, each accounting for about 20\% of the total. Tops, Service, and Socks are relatively less frequent and comprise roughly 10\% of the total. The most commonly discussed attributes in the Clothing domain, from most to least, are General, Quality, Size, and Looking, ranging from 15.3\% to 34.5\%. The remaining attributes such as Brand and Style have a much lower proportion.

\paragraph{Sentiment Distribution}

We further analyze the distribution of sentiment polarity in each domain, as shown in Table~\ref{tab:polarity_distribution}. The distribution of positive and negative sentiment polarity in each domain is generally imbalanced, with the Hotel domain being the most significant, where positive sentiment polarity accounts for over 90\%. The proportion of neutral sentiment polarity in each domain is relatively low, ranging from 0.3\% to 5.5\% of the total.

\section{Tasks and Evaluation}

On the basis of our dataset, we evaluate various baseline systems on six typical ABSA subtasks, including aspect extraction, aspect-based sentiment classification, aspect-opinion pair extraction, as well as aspect-opinion-sentiment triple extraction, aspect-category-sentiment triple extraction, aspect-category-opinion-sentiment quadruple extraction.

\subsection{Tasks with Baseline Systems}

The selected subtasks include both classification and extraction tasks. We choose different baseline systems for each subtask and compare their performance in the open domain setting as well as on the explicit and implicit test sets.

\paragraph{Aspect Extraction}

The task aims to extract explicit aspects from text, we select and modify baseline models based on both non-generative and generative architectures.

\textbf{Non-generative baselines:} \cite{devlin2018bert} proposed BERT, which has achieved good results on many natural language understanding tasks. We use BERT and BERT-CRF (which adds a CRF layer to BERT) as two basic non-generative baselines and replace BERT with BERT-XD (proposed by \cite{xu2019bert}) to form two new baselines, BERT-XD and BERT-XD+CRF.

\textbf{Generative baselines:} \cite{yan-etal-2021-unified} proposed a unified framework based on the generative model BART \cite{lewis2019bart} that can be used for various ABSA subtasks. We refer to this framework as BART-index and replace the generative model with T5 \cite{raffel2020exploring} to form a new baseline, T5-index. In addition, following \cite{zhang2021aspect}, we convert the true labels as text sequences corresponding to different templates and refer to this baseline as T5-Paraphrase.

\paragraph{Aspect-Based Sentiment Classification}

ABSC aims to determine the sentiment polarity of a given aspect. In addition to using BERT, BERT-XD, BART-index, and T5-index as baselines, we also use RoBERTa-MLP proposed by \cite{dai2021does} as a baseline, and refer to it as RoBERTaABSA.

\paragraph{Aspect-Sentiment Pair Extraction}

The goal of ASPE is to extract both the aspect and its corresponding sentiment polarity from the review text. When selecting baselines, in addition to BERT, BERT-XD, BERT-CRF, BERT-XD-CRF, BART-index, and T5-index mentioned earlier, we also design an output template for T5-Paraphrase as "<Aspect> is <Sentiment>". Additionally, we use the pipeline and joint methods proposed by \cite{hu2019open} as baselines, refer to as SpanABSA-pipeline and SpanABSA-joint, respectively.

\begin{table*}[!tp]
	\scriptsize

	\centering
	\resizebox{\linewidth}{!}{
	\setlength{\tabcolsep}{2mm}{
            \renewcommand{\arraystretch}{1.3}
		\begin{tabular} {l|cc|cc|cc|cc|cc}
		\hline
		& \multicolumn{2}{c|}{\textbf{Books}} &
            \multicolumn{2}{c|}{\textbf{Clothing}} & \multicolumn{2}{c|}{\textbf{Hotel}} & \multicolumn{2}{c|}{\textbf{Restaurant}} & \multicolumn{2}{c}{\textbf{Laptop}} \\
            
            & \multicolumn{1}{c}{\textbf{AE}} &
            \multicolumn{1}{c|}{\textbf{ASPE}} &

            \multicolumn{1}{c}{\textbf{AE}} &
            \multicolumn{1}{c|}{\textbf{ASPE}} &

            \multicolumn{1}{c}{\textbf{AE}} &
            \multicolumn{1}{c|}{\textbf{ASPE}} &

            \multicolumn{1}{c}{\textbf{AE}} &
            \multicolumn{1}{c|}{\textbf{ASPE}} &

            \multicolumn{1}{c}{\textbf{AE}} &
            \multicolumn{1}{c}{\textbf{ASPE}}

            \\
		\hline \hline
		BERT & 56.72 & 58.88 &  71.53 & 75.17 & 79.50 & 83.39 & 64.41 & 68.82 & 73.39 & 71.57 \\
		\hline
            BERT-XD & 58.38 & 60.43 & 73.89 & 78.03 & 79.69 & 83.91 & 66.18 & 72.40 & 75.45 & 74.72 \\
		\hline
            BERT+CRF & 61.23 & 59.70 &  74.46 & 76.35 & 81.27 & 84.11 & 67.03 & 70.00 & 75.70 & 72.91  \\
		\hline
            BERT-XD+CRF & 61.52 & 61.66 & 74.75 & 77.17 & 81.03 & \textbf{84.72} & 68.50 & 72.52 & 75.62 &  74.69 \\
		\hline
            SpanABSA-pipeline & - & 59.75 & - & 76.28 & - & 82.99 & - & 70.52 & - & 73.33 \\
		\hline
            SpanABSA-joint & - & 59.75 & - & 74.40 & - & 80.62 & - & 68.71 & - & 72.83  \\
		\hline \hline
            BART-index & 70.55 & 61.28 & 80.49 & 76.40 & 85.14 & 83.99 & 78.44 & 73.04 & 81.78 & 72.98  \\
		\hline
            T5-index & 72.46 & \textbf{65.77} & 83.17 & 78.93 & \textbf{85.74} & 84.15 & 78.80 & \textbf{75.48} & 82.42 & 75.58  \\
		\hline
            T5-Paraphrase & \textbf{72.52} & 65.05 & \textbf{83.53} & \textbf{79.56} & 85.54 & 84.43 & \textbf{80.10} & 74.67 & \textbf{83.10} & \textbf{75.98}  \\
		\hline

	\end{tabular}}}
	\caption{The micro-F1 results on Aspect Extraction (AE) and Aspect-Sentiment Pair Extraction (ASPE). The best results are in bold, each result is averaged over three runs with different random seeds.}
	\label{tab:explicit_ae_aspe_results}
\end{table*}

\begin{table*}[!tp]
	\scriptsize

	\centering
	\resizebox{\linewidth}{!}{
	\setlength{\tabcolsep}{2mm}{
            \renewcommand{\arraystretch}{1.3}
		\begin{tabular} {l|cc|cc|cc|cc|cc}
		\hline
		& \multicolumn{2}{c|}{\textbf{Books}} &
            \multicolumn{2}{c|}{\textbf{Clothing}} & \multicolumn{2}{c|}{\textbf{Hotel}} & \multicolumn{2}{c|}{\textbf{Restaurant}} & \multicolumn{2}{c}{\textbf{Laptop}} \\
            
            & \multicolumn{1}{c}{\textit{ACC}} &
            \multicolumn{1}{c|}{\textit{Macro-F1}} &

            \multicolumn{1}{c}{\textit{ACC}} &
            \multicolumn{1}{c|}{\textit{Macro-F1}} &

            \multicolumn{1}{c}{\textit{ACC}} &
            \multicolumn{1}{c|}{\textit{Macro-F1}} &

            \multicolumn{1}{c}{\textit{ACC}} &
            \multicolumn{1}{c|}{\textit{Macro-F1}} &

            \multicolumn{1}{c}{\textit{ACC}} &
            \multicolumn{1}{c}{\textit{Macro-F1}}

            \\
		\hline \hline
		BERT & 82.39 & 54.97 & 90.60 & 58.70 & 96.95 & 82.38 & 88.57 & 56.74 & 88.11 & 71.67 \\
		\hline
            BERT-XD & 83.07 & 54.54 & 91.56 & 60.16 & 97.13 & 84.12 & 89.50 & 61.31 & 88.71 & 65.45 \\
		\hline
            RoBERTaABSA & \textbf{88.33} & 61.77 & \textbf{93.83} & \textbf{62.00} & \textbf{97.30} & \textbf{87.24} & 91.89 & 63.30 & \textbf{90.93} & \textbf{76.49} \\
		\hline \hline
            BART-index & 83.48 & 54.97 & 89.22 & 57.78 & 97.16 & 86.18 & 90.89 & 71.34 & 89.50 & 72.67  \\
		\hline
            T5-index & 86.33 & \textbf{66.34} & 91.23 & 60.75 & 95.87 & 85.37 & \textbf{92.34} & \textbf{74.91} & 88.72 & 74.06 \\
		\hline

	\end{tabular}}}
	\caption{Results on the explicit Aspect-Based Sentiment Classification (ABSC) task.}
	\label{tab:explicit_absc_results}
\end{table*}

\paragraph{Aspect-Opinion-Sentiment Triple Extraction}

The goal of AOS triple extraction is to extract aspect-opinion-sentiment triples from the review text. We select non-generative and generative baselines for this subtask.
\textbf{Non-generative baselines}: \cite{wu2020grid} proposed a grid-based framework GTS, which can be used for joint extraction and classification tasks. \cite{chen2021bidirectional} transformed this task into a machine reading comprehension task and applied the Bidirectional MRC (BMRC) framework to solve the problem.
\textbf{Generative baselines}: 
In addition to BART-index, T5-index, and T5-Paraphrase, we use the GAS proposed by \cite{zhang2021towards} as a baseline, referred to as GAS (annotation) and GAS (extraction), respectively. We also replace the backbone of GAS with BART to obtain two baselines GAS-BART (annotation) and GAS-BART (extraction).
Considering that the aspect in this task cannot be implicit, we divide the test sets into two types for evaluation: the first test set is the original test set, and the second test set only contains triples with explicit aspects.

\paragraph{Aspect-Category-Sentiment Triple Extraction}

The goal of ACS triple extraction is to extract aspect-category-sentiment triples from review text \cite{wan2020target}. We select the generative model T5-Paraphrase, GAS (extraction), and GAS (annotation) as baselines and evaluate their performance on the aforementioned two types of test sets.

\paragraph{Aspect-Category-Opinion-Sentiment Quadruple Extraction}

The goal of ACOS quadruple extraction \cite{cai2021aspect,zhang2021aspect} is to extract the ACOS quadruples from the review text. We choose GAS-BART (Extraction), BART-Paraphrase, GAS (Extraction), T5-Paraphrase, BART-index, and T5-index adapted for implicit aspects as baselines. We evaluate the performance of these baselines on the two types of test sets mentioned above.

\subsection{Experimental Settings}

In order to enable the baseline systems to handle implicit examples, we make corresponding adjustments to baselines such as BART-index, T5-index, and T5-Paraphrase, which can only be used to handle explicit examples. For BART-index and T5-index, we define the implicit problem as an extraction problem, adding tokens "IA" and "IO" at the end of each sentence, representing implicit aspect and implicit opinion, respectively. For T5-Paraphrase, we use "it" to represent the implicit aspect and "implicitly expressed" to represent the implicit opinion and embed them into the generation template to generate tuples containing implicit aspect and opinion. In all models, we initialize the backbone with BERT-base, RoBERTa-base, BART-base, and T5-base parameters.

For each baseline, we select the average results of their performance on seeds 42, 153, and 650 as the final performance on the corresponding subtask. The original dataset was divided into training, validation, and test sets in a ratio of 7:1:2.

In the evaluation phase, for the extraction tasks, we use Micro-F1 as the evaluation metric and a triple is considered correct only if all elements of the triple are exactly the same as those in the gold triple. For the classification task, we use \textit{ACC} and \textit{Macro-F1} as the evaluation metrics.

\subsection{Main Results}

In this section, we evaluate and analyze the results of different baseline systems on each subtask.

\paragraph{Results on AE and ASPE}

As shown in Table~\ref{tab:explicit_ae_aspe_results}, the generative baselines T5-index and T5-Paraphrase achieve results higher than non-generative baselines in almost all domains, with a more pronounced advantage in the AE task. Meanwhile, in the ASPE task, it can be seen that the results are generally not lower than, and even higher than, those of the AE task in each domain, indicating that the ABSC task and AE task in non-generative architecture baselines can mutually promote each other, achieving better joint extraction performance while also improving the AE results.

\begin{table}[!tp]
	\large  
        % \tiny
	%	\renewcommand\tabcolsep{7.0pt} % ?????????
	\centering
	\resizebox{\linewidth}{!}{
        \renewcommand{\arraystretch}{1.3}
	\setlength{\tabcolsep}{2mm}{
		\begin{tabular} {l|ccccc}
			\hline
            & \multicolumn{1}{c}{\textbf{Books}} &
            \multicolumn{1}{c}{\textbf{Clothing}} &
            \multicolumn{1}{c}{\textbf{Hotel}}&
            \multicolumn{1}{c}{\textbf{Restaurant}}&
            \multicolumn{1}{c}{\textbf{Laptop}} \\
			\hline \hline
		GTS & 54.77 & 71.07 & 80.25 &  64.64 & 68.53  \\
			\hline
            BMRC & 57.85 & 70.70 & 81.10 & 65.14 & 71.39  \\
			\hline \hline
            GAS-BART (annotation) & 51.75 & 66.97 & 74.32 & 61.67 & 68.42 \\
			\hline
            GAS-BART (extraction) & 53.23 & 69.04 & 73.45 & 62.25 & 68.89 \\
			\hline
            BART-Paraphrase & 55.90 & 69.72 & 78.04 & 63.26 & 70.54 \\
			\hline
            BART-index & 57.23 & 71.99 &  79.66 & 67.22 & 70.22 \\
			\hline
            GAS (annotation) & 55.35 & 71.82 & 76.92 & 65.56 & 71.87 \\
			\hline
            GAS (extraction) & 56.99 & 71.97 & 75.61 & 66.10 & 72.14 \\
			\hline
            T5-Paraphrase & 58.68 & 73.56 & 79.21 & 67.15 & \textbf{72.18} \\
			\hline
            T5-index & \textbf{60.29} & \textbf{75.78} & \textbf{81.95} & \textbf{68.82} & 71.19 \\
			\hline
	\end{tabular}}}
	\caption{Results on aos triple extraction task.}
	\label{tab:exp_aos_results}
\end{table}

\begin{table}[!tp]
	\scriptsize
	\centering
	\resizebox{\linewidth}{!}{
        \renewcommand{\arraystretch}{1.3}
	\setlength{\tabcolsep}{2mm}{
		\begin{tabular} {l|ccccc}
			\hline
		
            & \multicolumn{1}{c}{\textbf{Books}} &
            \multicolumn{1}{c}{\textbf{Clothing}} &
            \multicolumn{1}{c}{\textbf{Hotel}}&
            \multicolumn{1}{c}{\textbf{Restaurant}}&
            \multicolumn{1}{c}{\textbf{Laptop}} \\
			\hline \hline
            GAS(annotation) & 52.49 & 43.72 & 66.79 & 64.55 & 46.09  \\
			\hline
            GAS(extraction) & \textbf{52.63} & 46.63 & 67.82 & 65.34 & \textbf{47.13} \\
			\hline
            T5-Paraphrase & 51.39 & \textbf{47.01} & \textbf{74.33} & \textbf{65.52} & 47.09  \\
			\hline
	\end{tabular}}}
	\caption{Results on acs triple extraction task.}
	\label{tab:exp_acs_results}
\end{table}

\paragraph{Results on ABSC}

As shown in Table~\ref{tab:explicit_absc_results}, RoBERTaABSA and T5-index both perform better than other baselines. Specifically, RoBERTaABSA achieves almost the best results in the Books, Clothing, Hotel, and Laptop domains, while T5-index achieves the best macro-average results in the Restaurant and Books domains, about 3\% to 5\% higher than the second-best baseline.
In addition, RoBERTaABSA had a lower Macro-F1 on Restaurant compared to T5-index, by about 10\%, which may be due to its poor performance in predicting a certain category, leading to large fluctuations in results under different seeds, which in turn affects the macro-average results.

\begin{table}[!tp]
	\normalsize
	\centering
	\resizebox{\linewidth}{!}{
        \renewcommand{\arraystretch}{1.3}
	\setlength{\tabcolsep}{2mm}{
		\begin{tabular} {l|ccccc}
			\hline
		
            & \multicolumn{1}{c}{\textbf{Books}} &
            \multicolumn{1}{c}{\textbf{Clothing}} &
            \multicolumn{1}{c}{\textbf{Hotel}}&
            \multicolumn{1}{c}{\textbf{Restaurant}}&
            \multicolumn{1}{c}{\textbf{Laptop}} \\
			\hline \hline
            GAS-BART(extraction) & 43.82 & 37.96 & 64.37 &  57.54 & 41.69 \\
			\hline
            BART-Parapharse & 44.87 & 39.99 & 68.45 &  58.21 & 43.03 \\
			\hline
            BART-index & 47.70 & 44.50 & 70.50 &  61.20 & 42.51 \\
			\hline
            GAS(extraction) & 47.17 & \textbf{45.28} & 65.75 & 60.31  & \textbf{45.18} \\
			\hline
            T5-Parapharse & 46.43 & 45.15 &  69.32 &  61.79 & 43.44 \\
			\hline
            T5-index & \textbf{52.70} & 42.58 & \textbf{73.52} &  \textbf{63.76} & 40.61 \\
			\hline
	\end{tabular}}}
	\caption{Results on the acos quadruple extraction task.}
	\label{tab:exp_acos_results}
\end{table}

\begin{table*}[!tp]
	\footnotesize
	\centering
	% \resizebox{\linewidth}{!}{
        \renewcommand{\arraystretch}{1.3}
	\setlength{\tabcolsep}{2mm}{
		\begin{tabular} {c|c|c|c|c|c|c}
			\hline
		\multirow{1}{*}{\textbf{Trained on $\to$}        }
            & \multirow{2}{*}{\textbf{Books}} &
            \multirow{2}{*}{\textbf{Clothing}} &
            \multirow{2}{*}{\textbf{Hotel}}&
            \multirow{2}{*}{\textbf{Restaurant}}&
            \multirow{2}{*}{\textbf{Laptop}} & \multirow{2}{*}{\textbf{The Rest}} \\
            \multirow{1}{*}{\textbf{Evaluated on $\downarrow$}} &  &  &  &  &  & 
 \\
			\hline \hline
            \textbf{Books} & {\color[HTML]{808080} 58.68} & 43.27 & 46.74 & 48.26 & 31.84 & 51.78 \\
			\hline
            \textbf{Clothing} & 52.17 & {\color[HTML]{808080} 73.56} & 56.85 & 64.43 & 26.15 & 60.39
 \\
			\hline
            \textbf{Hotel} & 62.41 & 62.51 & {\color[HTML]{808080} 79.21} & 70.09 & 36.84 & 72.60
 \\
			\hline
            \textbf{Restaurant} & 52.92 & 50.68 & 56.01 &  {\color[HTML]{808080} 67.15} & 36.21 & 61.47
 \\
			\hline
            \textbf{Laptop} & 55.05 & 56.57 & 56.88 & 58.67 & {\color[HTML]{808080} 72.18} & 61.77
 \\
			\hline
	\end{tabular}}
        % }
	\caption{Results of the open domain aos extraction task. The results in gray font are the results of in-domain experiments.}
	\label{tab:open_domain_aos_results}
\end{table*}

\begin{table*}[!tp]
	\scriptsize
	\centering
	\resizebox{\linewidth}{!}{
        \renewcommand{\arraystretch}{1.3}
	\setlength{\tabcolsep}{2mm}{
		\begin{tabular} {l|ccccc|ccccc}
			\hline
			& \multicolumn{5}{c|}{\textbf{AOS}} & \multicolumn{5}{c}{\textbf{ACOS}} \\
            & \multicolumn{1}{c}{\textbf{Books}} &
            \multicolumn{1}{c}{\textbf{Clothing}} &
            \multicolumn{1}{c}{\textbf{Hotel}}&
            \multicolumn{1}{c}{\textbf{Restaurant}}&
            \multicolumn{1}{c|}{\textbf{Laptop}}

            & \multicolumn{1}{c}{\textbf{Books}} &
            \multicolumn{1}{c}{\textbf{Clothing}} &
            \multicolumn{1}{c}{\textbf{Hotel}}&
            \multicolumn{1}{c}{\textbf{Restaurant}}&
            \multicolumn{1}{c}{\textbf{Laptop}} \\
			\hline \hline

            GAS-BART(extraction) & 47.1 & 64.32 & 71.60 & 57.48 &  63.74 & 36.60 & 34.29 & 64.00 & 51.08 & 40.60 \\
			\hline
            BART-Parapharse & 50.56 & 64.60 & 75.18 & 59.23 & 65.31 & 37.00 & 37.00 & 67.17 & 52.47 & 41.60 \\
			\hline
            BART-index & 50.06 & 67.32 & 77.27 & 63.26 & 65.80 & 36.99 & 37.41 & 65.42 & 54.28 & 39.53 \\
			\hline
            GAS(extraction) & 50.79 & 66.52 & 72.59 & 62.95 & 67.22 & 38.94 & 40.20 & 64.44 & 54.20 & 43.15 \\
			\hline
            T5-Parapharse & 52.00 & 68.45 & 75.84 & 63.07 & \textbf{67.88} & 39.00 & \textbf{41.26} & 66.89 & 56.31 & \textbf{43.57} \\
			\hline
            T5-index & \textbf{53.46} & \textbf{69.63} & \textbf{77.98} & \textbf{65.12} & 66.06 & \textbf{41.71} & 37.55 & \textbf{69.22} & \textbf{56.49} & 	40.38 \\
			\hline
	\end{tabular}}}
	\caption{Results of the aos and acos tasks on datasets containing implicit samples.}
	\label{tab:exp_imp_aos_acos_results}
\end{table*}

\paragraph{Results on AOS Triple Extraction}

In Table~\ref{tab:exp_aos_results}, the generative method T5-index achieves almost the best results, about 1\% to 2\% higher than the second-best baseline in each domain. In the Hotel domain, the result of extracting the aos triple was around 80\%, which indicates that the T5-index can extract aspect-opinion pairs well in this domain, partly due to the sentiment imbalance problem of the Hotel domain. 
However, T5-index has lower results in the Books domain. On the one hand, the aspect extraction in this domain is more difficult compared to other domains, with an F1 score about 10\% lower than other domains. On the other hand, opinions related to aspects may also be difficult to extract, leading to relatively lower results in extracting the aos triples.

\paragraph{Results on ACS Triple Extraction}

In this task, we compare the results of three generative models, as shown in Table~\ref{tab:exp_acs_results}. T5-Paraphrase achieves the relatively best results in five domains, especially in the Hotel domain. Combined with the results shown in Table~\ref{tab:exp_aos_results}, we can speculate that compared to other baselines, T5-Paraphrase can better identify the category of the Hotel domain and achieve about 7\% improvement in extracting acs triples. In addition, GAS (extraction) has some advantages over GAS (annotation), reflecting the influence of output templates on the acs extraction task.

\paragraph{Results on ACOS Quadruple Extraction}

In the ACOS task, we compare the results of six generative models, as shown in Table~\ref{tab:exp_acos_results}. T5-index achieves the best results in the Books, Hotel, and Restaurant domains, with improvements of 5.00\%, 3.02\%, and 1.97\% over the second-best models, respectively. In addition, in the Hotel domain, the results of extracting acos quadruples can exceed 70\%, but in the Laptop domain, it can only reach about 40\%. This may be due to the fact that the Laptop domain has more categories, and the acos quadruples in the Hotel domain may be easier to identify compared to other domains.

\subsection{Cross-Domain and Out-of-Domain Evaluation}

In addition to the in-domain setting, we analyze the results of baseline models in cross-domain and out-of-domain settings. We select T5-Paraphrase, which performs well, as the baseline and conduct experiments on the aos task.

As shown in Table~\ref{tab:open_domain_aos_results}, under the cross-domain setting, the results are lower than those in the in-domain setting, with differences ranging from 10\% to 40\% when the source domain and target domain were inconsistent. Specifically, when the restaurant domain is the source domain, it has the best generalization ability compared to using other domains as the source domain.
On the other hand, when the laptop domain is the source domain and other domains are the target domain, the results of cross-domain training are relatively low, ranging from 26\% to 36\%, with a difference of up to 40\% compared to the results obtained when the laptop domain is the target domain, indicating poor transferability of the model and significant differences between the laptop domain and other domains.

We set the out-of-domain setting to combine the four domains into the source domain and use the remaining domain as the target domain. Specifically, we combine the training and validation sets of the other four domains as the new training and validation sets and test the final results on the test set of the remaining domain. It can be seen that except for the clothing domain, the experimental results of all other domains in the out-of-domain setting are better than those in the cross-domain setting, by about 2\% to 5\%, reflecting the importance of multi-source data in improving the performance of cross-domain model training.

\subsection{Implicit Evaluation}

We compare the results of different baselines on the aos task that contains implicit opinions, and the acos task that contains implicit aspects and opinions.
As shown in Table~\ref{tab:exp_imp_aos_acos_results}, T5-index and T5-paraphrase achieve almost optimal results in each domain, with T5-index performing better on average in the aos task than T5-paraphrase, and performing similarly to T5-paraphrase on average in the acos task.
Compared with the results in Table~\ref{tab:exp_aos_results} and Table~\ref{tab:exp_acos_results}, it can be seen that the results of training and testing with implicit examples are lower than those with only explicit examples in all domains. This indicates that it is more difficult to extract aos triples and acos quadruples that contain implicit examples compared to explicit ones.

\section{Related Work}

Aspect-based sentiment analysis (ABSA) has attracted wide attention from researchers in recent years, mainly focusing on the joint extraction of different elements.

\paragraph{Two-Element Extraction Tasks}

The two-element extraction tasks generally include Aspect-Sentiment Pair Extraction (ASPE) and Aspect-Opinion Pair Extraction (AOPE). Both tasks aim to jointly extract aspects and their related opinions, but the difference lies in the form of the opinion. The ASPE aims to identify sentiment polarity, while the AOPE aims to extract opinion expressions.
\cite{mitchell2013open} proposed three frameworks to model the ASPE task in an open domain setting, and many deep learning-based pipelines \cite{hu2019open} or end-to-end methods \cite{li2019unified,he2019interactive,chen2020relation,yan-etal-2021-unified} have been proposed to solve this task. 
For the AOPE task, \cite{hu2004mining} attempted to solve it with a pipeline approach, and in recent years, many deep learning-based works have been proposed to address this problem\cite{chen2020synchronous,zhao2020spanmlt,gao2021question,liu2022pair}.

\paragraph{Three-Element Extraction Tasks}

The three-element extraction tasks include Aspect-Opinion-Sentiment (AOS) triple extraction \cite{peng2020knowing} and Aspect-Category-Sentiment (ACS) triple extraction \cite{wan2020target}. Most existing research has focused on explicit AOS triple extraction task \cite{xu2020position,wu2020grid,mao2021joint,chen2021bidirectional,chen2022enhanced,zhai2022mrc}. As for the ACS triple extraction task, some generative methods \cite{zhang2021towards} have attempted to solve this problem with a specially designed input template.

\paragraph{Four-Element Extraction Task}

\cite{cai2021aspect} introduced implicit aspect and opinion into the task definition and proposed a four-element extraction task with both explicit and implicit aspect and opinion annotations. \cite{zhang2021aspect} also studied the explicit four-element extraction task and proposed a generative framework to extract the quadruples. Subsequently, some research works \cite{mao2022seq2path,hu2022improving,hu2023uncertainty} proposed methods to solve the ACOS extraction problem.

\paragraph{ABSA Datasets}

SemEval introduced benchmark datasets for ABSA (Aspect-Based Sentiment Analysis) \cite{pontiki-etal-2014-semeval,pontiki2015semeval,pontiki2016semeval}, which mainly include single elements (aspect, category, and sentiment), and aspect-sentiment pairs, aspect-category-sentiment triples on some datasets.
\cite{wang2016recursive,wang2017coupled} annotated opinions for the SemEval 2014 laptop and restaurant domains, as well as the 2015 restaurant domain, and \cite{chen2020synchronous} annotated aspect-opinion pairs based on their annotated opinion.
On the other hand, \cite{fan2019target} annotated aspect-opinion pairs for the SemEval 2014 laptop and restaurant domains, as well as the SemEval 2015 and 2016 restaurant domains. Subsequently, \cite{peng2020knowing,xu2020position} proposed the AOS datasets based on \cite{fan2019target} and SemEval datasets, \cite{xu2023measuring,chia2023domain} further supplemented AOS datasets for domains other than restaurant and laptop.
\cite{cai2021aspect} proposed the ACOS quadruple dataset for the restaurant and laptop domains, incorporating implicit aspects and opinions into the quadruple extraction task.
In contrast to the above datasets, the dataset we propose covers five different domains and can better evaluate the performance of models in various typical subtasks under an open-domain setting.

\section{Conclusion}

In the context of the emergence of large language models, ABSA remains a challenging problem. We propose a manually annotated ABSA dataset that contains quadruples, explicit and implicit aspects and opinions, and covers five domains. The scale of annotation is also much larger than that of previous datasets. We summarize and develop annotation guidelines for the dataset, and conduct comprehensive statistics and analysis on the annotated dataset. Meanwhile, under the open-domain setting, we organize evaluations of generative and non-generative baselines on several typical ABSA subtasks under in-domain, cross-domain, and out-of-domain settings.

% \input{Multi-domain ABSA/intro}

\bibliography{acl2021}

\begin{thebibliography}{46}
\expandafter\ifx\csname natexlab\endcsname\relax\def\natexlab#1{#1}\fi

\bibitem[{{\'A}lvarez-L{\'o}pez et~al.(2018){\'A}lvarez-L{\'o}pez,
  Fern{\'a}ndez-Gavilanes, Costa-Montenegro, and Bellot}]{alvarez2018proposal}
Tamara {\'A}lvarez-L{\'o}pez, Milagros Fern{\'a}ndez-Gavilanes, Enrique
  Costa-Montenegro, and Patrice Bellot. 2018.
\newblock A proposal for book oriented aspect based sentiment analysis:
  Comparison over domains.
\newblock In \emph{Natural Language Processing and Information Systems: 23rd
  International Conference on Applications of Natural Language to Information
  Systems, NLDB 2018, Paris, France, June 13-15, 2018, Proceedings 23}, pages
  3--14. Springer.

\bibitem[{Cai et~al.(2021)Cai, Xia, and Yu}]{cai2021aspect}
Hongjie Cai, Rui Xia, and Jianfei Yu. 2021.
\newblock Aspect-category-opinion-sentiment quadruple extraction with implicit
  aspects and opinions.
\newblock In \emph{Proceedings of the 59th Annual Meeting of the Association
  for Computational Linguistics and the 11th International Joint Conference on
  Natural Language Processing (ACL-IJCNLP)}, pages 340--350.

\bibitem[{Chen et~al.(2022)Chen, Zhai, Feng, Li, and Wang}]{chen2022enhanced}
Hao Chen, Zepeng Zhai, Fangxiang Feng, Ruifan Li, and Xiaojie Wang. 2022.
\newblock Enhanced multi-channel graph convolutional network for aspect
  sentiment triplet extraction.
\newblock In \emph{Proceedings of the 60th Annual Meeting of the Association
  for Computational Linguistics (Volume 1: Long Papers)}, pages 2974--2985.

\bibitem[{Chen et~al.(2020)Chen, Liu, Wang, Zhang, and
  Chi}]{chen2020synchronous}
Shaowei Chen, Jie Liu, Yu~Wang, Wenzheng Zhang, and Ziming Chi. 2020.
\newblock Synchronous double-channel recurrent network for aspect-opinion pair
  extraction.
\newblock In \emph{Proceedings of the 58th Annual Meeting of the Association
  for Computational Linguistics (ACL)}, pages 6515--6524.

\bibitem[{Chen et~al.(2021)Chen, Wang, Liu, and Wang}]{chen2021bidirectional}
Shaowei Chen, Yu~Wang, Jie Liu, and Yuelin Wang. 2021.
\newblock Bidirectional machine reading comprehension for aspect sentiment
  triplet extraction.
\newblock In \emph{Proceedings of the 35th AAAI Conference on Artificial
  Intelligence (AAAI)}, pages 12666--12674.

\bibitem[{Chen and Qian(2020)}]{chen2020relation}
Zhuang Chen and Tieyun Qian. 2020.
\newblock Relation-aware collaborative learning for unified aspect-based
  sentiment analysis.
\newblock In \emph{Proceedings of the 58th Annual Meeting of the Association
  for Computational Linguistics (ACL)}, pages 3685--3694.

\bibitem[{Chia et~al.(2023)Chia, Chen, Han, Chen, Aljunied, Poria, and
  Bing}]{chia2023domain}
Yew~Ken Chia, Hui Chen, Wei Han, Guizhen Chen, Sharifah~Mahani Aljunied,
  Soujanya Poria, and Lidong Bing. 2023.
\newblock Domain-expanded aste: Rethinking generalization in aspect sentiment
  triplet extraction.
\newblock \emph{arXiv preprint arXiv:2305.14434}.

\bibitem[{Dai et~al.(2021)Dai, Yan, Sun, Liu, and Qiu}]{dai2021does}
Junqi Dai, Hang Yan, Tianxiang Sun, Pengfei Liu, and Xipeng Qiu. 2021.
\newblock Does syntax matter? a strong baseline for aspect-based sentiment
  analysis with roberta.
\newblock \emph{arXiv preprint arXiv:2104.04986}.

\bibitem[{Devlin et~al.(2018)Devlin, Chang, Lee, and
  Toutanova}]{devlin2018bert}
Jacob Devlin, Ming-Wei Chang, Kenton Lee, and Kristina Toutanova. 2018.
\newblock Bert: Pre-training of deep bidirectional transformers for language
  understanding.
\newblock \emph{arXiv preprint arXiv:1810.04805}.

\bibitem[{Fan et~al.(2019)Fan, Wu, Dai, Huang, and Chen}]{fan2019target}
Zhifang Fan, Zhen Wu, Xinyu Dai, Shujian Huang, and Jiajun Chen. 2019.
\newblock Target-oriented opinion words extraction with target-fused neural
  sequence labeling.
\newblock In \emph{Proceedings of the 2019 Conference of the North American
  Chapter of the Association for Computational Linguistics (NAACL)}, pages
  2509--2518.

\bibitem[{Gao et~al.(2021)Gao, Wang, Liu, Wang, Zhang, and
  Liao}]{gao2021question}
Lei Gao, Yulong Wang, Tongcun Liu, Jingyu Wang, Lei Zhang, and Jianxin Liao.
  2021.
\newblock Question-driven span labeling model for aspect-opinion pair
  extraction.
\newblock In \emph{Proceedings of the AAAI conference on artificial
  intelligence (AAAI)}, pages 12875--12883.

\bibitem[{He et~al.(2019)He, Lee, Ng, and Dahlmeier}]{he2019interactive}
Ruidan He, Wee~Sun Lee, Hwee~Tou Ng, and Daniel Dahlmeier. 2019.
\newblock An interactive multi-task learning network for end-to-end
  aspect-based sentiment analysis.
\newblock In \emph{Proceedings of the 57th Annual Meeting of the Association
  for Computational Linguistics (ACL)}, pages 504--515.

\bibitem[{Hu et~al.(2023)Hu, Bai, Wu, Zhang, Zhang, Gao, Zhao, and
  Huang}]{hu2023uncertainty}
Mengting Hu, Yinhao Bai, Yike Wu, Zhen Zhang, Liqi Zhang, Hang Gao, Shiwan
  Zhao, and Minlie Huang. 2023.
\newblock Uncertainty-aware unlikelihood learning improves generative aspect
  sentiment quad prediction.
\newblock \emph{arXiv preprint arXiv:2306.00418}.

\bibitem[{Hu et~al.(2022)Hu, Wu, Gao, Bai, and Zhao}]{hu2022improving}
Mengting Hu, Yike Wu, Hang Gao, Yinhao Bai, and Shiwan Zhao. 2022.
\newblock Improving aspect sentiment quad prediction via template-order data
  augmentation.
\newblock \emph{arXiv preprint arXiv:2210.10291}.

\bibitem[{Hu et~al.(2019)Hu, Peng, Huang, Li, and Lv}]{hu2019open}
Minghao Hu, Yuxing Peng, Zhen Huang, Dongsheng Li, and Yiwei Lv. 2019.
\newblock Open-domain targeted sentiment analysis via span-based extraction and
  classification.
\newblock In \emph{Proceedings of the 57th Annual Meeting of the Association
  for Computational Linguistics (ACL)}, pages 537--546.

\bibitem[{Hu and Liu(2004)}]{hu2004mining}
Minqing Hu and Bing Liu. 2004.
\newblock Mining and summarizing customer reviews.
\newblock In \emph{Proceedings of the tenth ACM SIGKDD international conference
  on Knowledge discovery and data mining}, pages 168--177.

\bibitem[{Klie et~al.(2018)Klie, Bugert, Boullosa, de~Castilho, and
  Gurevych}]{klie2018inception}
Jan-Christoph Klie, Michael Bugert, Beto Boullosa, Richard~Eckart de~Castilho,
  and Iryna Gurevych. 2018.
\newblock The inception platform: Machine-assisted and knowledge-oriented
  interactive annotation.
\newblock In \emph{proceedings of the 27th international conference on
  computational linguistics: system demonstrations}, pages 5--9.

\bibitem[{Lewis et~al.(2019)Lewis, Liu, Goyal, Ghazvininejad, Mohamed, Levy,
  Stoyanov, and Zettlemoyer}]{lewis2019bart}
Mike Lewis, Yinhan Liu, Naman Goyal, Marjan Ghazvininejad, Abdelrahman Mohamed,
  Omer Levy, Ves Stoyanov, and Luke Zettlemoyer. 2019.
\newblock Bart: Denoising sequence-to-sequence pre-training for natural
  language generation, translation, and comprehension.
\newblock \emph{arXiv preprint arXiv:1910.13461}.

\bibitem[{Li et~al.(2019)Li, Bing, Li, and Lam}]{li2019unified}
Xin Li, Lidong Bing, Piji Li, and Wai Lam. 2019.
\newblock A unified model for opinion target extraction and target sentiment
  prediction.
\newblock In \emph{Proceedings of the 33rd AAAI Conference on Artificial
  Intelligence (AAAI)}, pages 6714--6721.

\bibitem[{Liu(2012)}]{liu2012sentiment}
Bing Liu. 2012.
\newblock Sentiment analysis and opinion mining.
\newblock \emph{Synthesis lectures on human language technologies}, pages
  1--167.

\bibitem[{Liu et~al.(2022)Liu, Li, Fei, and Ji}]{liu2022pair}
Yijiang Liu, Fei Li, Hao Fei, and Donghong Ji. 2022.
\newblock Pair-wise aspect and opinion terms extraction as graph parsing via a
  novel mutually-aware interaction mechanism.
\newblock \emph{Neurocomputing}, 493.

\bibitem[{Mao et~al.(2022)Mao, Shen, Yang, Zhu, and Cai}]{mao2022seq2path}
Yue Mao, Yi~Shen, Jingchao Yang, Xiaoying Zhu, and Longjun Cai. 2022.
\newblock Seq2path: Generating sentiment tuples as paths of a tree.
\newblock In \emph{Findings of the Association for Computational Linguistics:
  ACL 2022}, pages 2215--2225.

\bibitem[{Mao et~al.(2021)Mao, Shen, Yu, and Cai}]{mao2021joint}
Yue Mao, Yi~Shen, Chao Yu, and Longjun Cai. 2021.
\newblock A joint training dual-mrc framework for aspect based sentiment
  analysis.
\newblock \emph{arXiv preprint arXiv:2101.00816}.

\bibitem[{Mitchell et~al.(2013)Mitchell, Aguilar, Wilson, and
  Van~Durme}]{mitchell2013open}
Margaret Mitchell, Jacqui Aguilar, Theresa Wilson, and Benjamin Van~Durme.
  2013.
\newblock Open domain targeted sentiment.
\newblock In \emph{Proceedings of the 2013 Conference on Empirical Methods in
  Natural Language Processing (EMNLP)}, pages 1643--1654.

\bibitem[{Ni et~al.(2019)Ni, Li, and McAuley}]{ni2019justifying}
Jianmo Ni, Jiacheng Li, and Julian McAuley. 2019.
\newblock Justifying recommendations using distantly-labeled reviews and
  fine-grained aspects.
\newblock In \emph{Proceedings of the 2019 conference on empirical methods in
  natural language processing and the 9th international joint conference on
  natural language processing (EMNLP-IJCNLP)}, pages 188--197.

\bibitem[{Peng et~al.(2020)Peng, Xu, Bing, Huang, Lu, and Si}]{peng2020knowing}
Haiyun Peng, Lu~Xu, Lidong Bing, Fei Huang, Wei Lu, and Luo Si. 2020.
\newblock Knowing what, how and why: A near complete solution for aspect-based
  sentiment analysis.
\newblock In \emph{Proceedings of the 34th AAAI Conference on Artificial
  Intelligence (AAAI)}, pages 8600--8607.

\bibitem[{Pontiki et~al.(2016)Pontiki, Galanis, Papageorgiou, Androutsopoulos,
  Manandhar, Al-Smadi, Al-Ayyoub, Zhao, Qin, De~Clercq
  et~al.}]{pontiki2016semeval}
Maria Pontiki, Dimitrios Galanis, Haris Papageorgiou, Ion Androutsopoulos,
  Suresh Manandhar, Mohammad Al-Smadi, Mahmoud Al-Ayyoub, Yanyan Zhao, Bing
  Qin, Orph{\'e}e De~Clercq, et~al. 2016.
\newblock Semeval-2016 task 5: Aspect based sentiment analysis.
\newblock In \emph{International workshop on semantic evaluation}, pages
  19--30.

\bibitem[{Pontiki et~al.(2015)Pontiki, Galanis, Papageorgiou, Manandhar, and
  Androutsopoulos}]{pontiki2015semeval}
Maria Pontiki, Dimitrios Galanis, Harris Papageorgiou, Suresh Manandhar, and
  Ion Androutsopoulos. 2015.
\newblock Semeval-2015 task 12: Aspect based sentiment analysis.
\newblock In \emph{Proceedings of the 9th international workshop on semantic
  evaluation (SemEval 2015)}, pages 486--495.

\bibitem[{Pontiki et~al.(2014)Pontiki, Galanis, Pavlopoulos, Papageorgiou,
  Androutsopoulos, and Manandhar}]{pontiki-etal-2014-semeval}
Maria Pontiki, Dimitris Galanis, John Pavlopoulos, Harris Papageorgiou, Ion
  Androutsopoulos, and Suresh Manandhar. 2014.
\newblock {S}em{E}val-2014 task 4: Aspect based sentiment analysis.
\newblock In \emph{Proceedings of the 8th International Workshop on Semantic
  Evaluation ({S}em{E}val 2014)}, pages 27--35, Dublin, Ireland. Association
  for Computational Linguistics.

\bibitem[{Raffel et~al.(2020)Raffel, Shazeer, Roberts, Lee, Narang, Matena,
  Zhou, Li, and Liu}]{raffel2020exploring}
Colin Raffel, Noam Shazeer, Adam Roberts, Katherine Lee, Sharan Narang, Michael
  Matena, Yanqi Zhou, Wei Li, and Peter~J Liu. 2020.
\newblock Exploring the limits of transfer learning with a unified text-to-text
  transformer.
\newblock \emph{The Journal of Machine Learning Research}, 21(1):5485--5551.

\bibitem[{Wan et~al.(2020)Wan, Yang, Du, Liu, Qi, and Pan}]{wan2020target}
Hai Wan, Yufei Yang, Jianfeng Du, Yanan Liu, Kunxun Qi, and Jeff~Z Pan. 2020.
\newblock Target-aspect-sentiment joint detection for aspect-based sentiment
  analysis.
\newblock In \emph{Proceedings of the 34th AAAI Conference on Artificial
  Intelligence (AAAI)}, pages 9122--9129.

\bibitem[{Wang et~al.(2016)Wang, Pan, Dahlmeier, and Xiao}]{wang2016recursive}
Wenya Wang, Sinno~Jialin Pan, Daniel Dahlmeier, and Xiaokui Xiao. 2016.
\newblock Recursive neural conditional random fields for aspect-based sentiment
  analysis.
\newblock \emph{arXiv preprint arXiv:1603.06679}.

\bibitem[{Wang et~al.(2017)Wang, Pan, Dahlmeier, and Xiao}]{wang2017coupled}
Wenya Wang, Sinno~Jialin Pan, Daniel Dahlmeier, and Xiaokui Xiao. 2017.
\newblock Coupled multi-layer attentions for co-extraction of aspect and
  opinion terms.
\newblock In \emph{Proceedings of the 31st AAAI Conference on Artificial
  Intelligence (AAAI)}, pages 3316--3322.

\bibitem[{Wang et~al.(2023)Wang, Xie, Ding, Feng, and Xia}]{wang2023chatgpt}
Zengzhi Wang, Qiming Xie, Zixiang Ding, Yi~Feng, and Rui Xia. 2023.
\newblock Is chatgpt a good sentiment analyzer? a preliminary study.
\newblock \emph{arXiv preprint arXiv:2304.04339}.

\bibitem[{Wilson et~al.(2005)Wilson, Wiebe, and
  Hoffmann}]{wilson2005recognizing}
Theresa Wilson, Janyce Wiebe, and Paul Hoffmann. 2005.
\newblock Recognizing contextual polarity in phrase-level sentiment analysis.
\newblock In \emph{Proceedings of human language technology conference and
  conference on empirical methods in natural language processing}, pages
  347--354.

\bibitem[{Wu et~al.(2020)Wu, Ying, Zhao, Fan, Dai, and Xia}]{wu2020grid}
Zhen Wu, Chengcan Ying, Fei Zhao, Zhifang Fan, Xinyu Dai, and Rui Xia. 2020.
\newblock Grid tagging scheme for end-to-end fine-grained opinion extraction.
\newblock In \emph{Proceedings of the 2020 Conference on Empirical Methods in
  Natural Language Processing: Findings}, pages 2576--2585.

\bibitem[{Xu et~al.(2019)Xu, Liu, Shu, and Philip}]{xu2019bert}
Hu~Xu, Bing Liu, Lei Shu, and S~Yu Philip. 2019.
\newblock Bert post-training for review reading comprehension and aspect-based
  sentiment analysis.
\newblock In \emph{Proceedings of the 2019 Conference of the North American
  Chapter of the Association for Computational Linguistics (NAACL)}, pages
  2324--2335.

\bibitem[{Xu et~al.(2020)Xu, Li, Lu, and Bing}]{xu2020position}
Lu~Xu, Hao Li, Wei Lu, and Lidong Bing. 2020.
\newblock Position-aware tagging for aspect sentiment triplet extraction.
\newblock \emph{arXiv preprint arXiv:2010.02609}.

\bibitem[{Xu et~al.(2023)Xu, Yang, Wu, Chen, Zhao, and Dai}]{xu2023measuring}
Ting Xu, Huiyun Yang, Zhen Wu, Jiaze Chen, Fei Zhao, and Xinyu Dai. 2023.
\newblock Measuring your aste models in the wild: A diversified multi-domain
  dataset for aspect sentiment triplet extraction.
\newblock \emph{arXiv preprint arXiv:2305.17448}.

\bibitem[{Yan et~al.(2021)Yan, Dai, Ji, Qiu, and Zhang}]{yan-etal-2021-unified}
Hang Yan, Junqi Dai, Tuo Ji, Xipeng Qiu, and Zheng Zhang. 2021.
\newblock \href {https://aclanthology.org/2021.acl-long.188} {A unified
  generative framework for aspect-based sentiment analysis}.
\newblock In \emph{Proceedings of the 59th Annual Meeting of the Association
  for Computational Linguistics and the 11th International Joint Conference on
  Natural Language Processing (Volume 1: Long Papers)}, pages 2416--2429.

\bibitem[{Zhai et~al.(2022)Zhai, Chen, Feng, Li, and Wang}]{zhai2022mrc}
Zepeng Zhai, Hao Chen, Fangxiang Feng, Ruifan Li, and Xiaojie Wang. 2022.
\newblock Com-mrc: A context-masked machine reading comprehension framework for
  aspect sentiment triplet extraction.
\newblock In \emph{Proceedings of the 2022 Conference on Empirical Methods in
  Natural Language Processing (EMNLP)}, pages 3230--3241.

\bibitem[{Zhang et~al.(2021{\natexlab{a}})Zhang, Deng, Li, Yuan, Bing, and
  Lam}]{zhang2021aspect}
Wenxuan Zhang, Yang Deng, Xin Li, Yifei Yuan, Lidong Bing, and Wai Lam.
  2021{\natexlab{a}}.
\newblock Aspect sentiment quad prediction as paraphrase generation.
\newblock In \emph{Proceedings of the 2021 Conference on Empirical Methods in
  Natural Language Processing (EMNLP)}, pages 9209--9219.

\bibitem[{Zhang et~al.(2021{\natexlab{b}})Zhang, Li, Deng, Bing, and
  Lam}]{zhang2021towards}
Wenxuan Zhang, Xin Li, Yang Deng, Lidong Bing, and Wai Lam. 2021{\natexlab{b}}.
\newblock Towards generative aspect-based sentiment analysis.
\newblock In \emph{Proceedings of the 59th Annual Meeting of the Association
  for Computational Linguistics and the 11th International Joint Conference on
  Natural Language Processing (Volume 2: Short Papers)}, pages 504--510.

\bibitem[{Zhao et~al.(2020)Zhao, Huang, Zhang, Lu et~al.}]{zhao2020spanmlt}
He~Zhao, Longtao Huang, Rong Zhang, Quan Lu, et~al. 2020.
\newblock Spanmlt: A span-based multi-task learning framework for pair-wise
  aspect and opinion terms extraction.
\newblock In \emph{Proceedings of the 58th Annual Meeting of the Association
  for Computational Linguistics (ACL)}, pages 3239--3248.

\bibitem[{Zhao et~al.(2014)Zhao, Qin, and Liu}]{zhao2014creating}
Yanyan Zhao, Bing Qin, and Ting Liu. 2014.
\newblock Creating a fine-grained corpus for chinese sentiment analysis.
\newblock \emph{IEEE Intelligent Systems}, pages 36--43.

\bibitem[{Zong et~al.(2021)Zong, Xia, and Zhang}]{zong2021text}
Chengqing Zong, Rui Xia, and Jiajun Zhang. 2021.
\newblock \emph{Text data mining}, volume 711.
\newblock Springer.

\end{thebibliography}
\bibliographystyle{acl_natbib}

\appendix

% \section{What To Do Next?}

% 1. document-level ABSA

% 2. enhancing coreference resolution

% 3. 

\section{Annotation Guidelines}
\label{appendix:annotation}

In this section, we provide annotated examples, annotation rules for each element, and define categories for the Clothing domain. During the annotation process, the annotators need to identify the aspect with opinion expression and its corresponding category, opinion, and sentiment in the review text. The annotation process for each annotator is as follows:

1. Read the current sentence and determine whether it belongs to the assigned annotation domain. If it does, go to step 2. If not, select N/A at the end of the sentence and go to step 3.

2. Determine whether there is an explicit or implicit aspect with opinion expression in the sentence (determined by the aspect and category, if not, go to step 3), and annotate the corresponding aspect-category-opinion-sentiment quadruple. Then, go back to step 2.

3. End the annotation of the current sentence and continue to annotate the next sentence until the specified number of sentences is reached.

\subsection{Annotation Examples}
This section provides annotated examples of quadruples containing explicit and implicit aspects and opinions.

\textbf{Example 1.1}:
The screen is good, but the backlit keyboard is awful.

Quadruples:

screen-Display\#General-good-positive

backlit keyboard-Keyboard\#General-awful-Negative

\textbf{Example 1.2}:
I had a great experience.

Quadruples:

NULL-Restaurant\#General-great-positive

Explanation: The aspect in the sentence is implicit, and we use NULL to represent the aspect, and the category Restaurant\#General to represent the opinion target.

\textbf{Example 1.3}:
The app takes seconds to respond.

Quadruples:

app-Software\#General-NULL-negative

Explanation: The sentiment polarity of the sentence towards the "app" is negative, but there is no explicit opinion. Here we use NULL to represent the opinion.

\textbf{Example 1.4}:
It takes three hours to have the seat.

Quadruples:

NULL-Restaurant\#General-NULL-negative

Explanation: The aspect and opinion in the sentence are implicit, but it expresses negative sentiment polarity toward the restaurant.

\subsection{Aspect Annotation Examples}

When it is difficult to determine the annotation of the aspect, we determine whether the aspect is explicit or implicit by judging whether it can correspond to a predefined entity in the category and whether it is directly related to the domain.

\textbf{Example 2.1}:
It is a great purchase.

Quadruples:

NULL-Book\#General-great-positive

Explanation: "Purchase" is the action of the commenter and is not directly related to the domain (Book). It cannot be found in any corresponding entity, so it is marked as an implicit aspect.

\textbf{Example 2.2}:
A good stay.

Quadruples:

NULL-Hotel\#General-good-positive

Explanation: Similarly, "stay" is not directly related to the topic and cannot be found in any corresponding entity. Therefore, it is marked as an implicit aspect.

Below are some examples of the principles for annotating the boundaries of the aspect.

\textbf{Example 2.3}:
The Chinese restaurant is great, The Italian restaurant nearby is also good.

Quadruples:

Chinese restaurant-Restaurant\#General-great-positive

Italian restaurant-Restaurant\#General-good-positive

Explanation: "Chinese restaurant" and "Italian restaurant" correspond to objective things, so we cannot just annotate "restaurant" as aspects. Also, the definite article "the" does not need to be included in the scope.

\subsection{Opinion Annotation Examples}

In the annotation of opinions, only the opinion words (including adjectives, adverbs, verbs, and a few nouns) need to be annotated.

\textbf{Example 3.1}:
The various stories flow well.

Quadruples:

various stories-Content\#General-well-positive

Explanation: In this example, only "well" needs to be annotated as opinion, and there is no need to also annotate the "flow" describing the story's progression.

Annotation of sentiment intensifiers, diminishers, and negation:

\textbf{Example 3.2}:
Very lovely laptop.

Quadruples:

laptop-laptop\#General-Very lovely-positive

Explanation: In this example, "very" expresses the intensify of the positive sentiment conveyed by "lovely", and needs to be annotated in the opinion.

\textbf{Example 3.3}:
Service was kind of slow.

Quadruples:

Service-Service\#General-kind of slow-negative

Explanation: "kind of" expresses the diminish of the negative sentiment and needs to be annotated.

\textbf{Example 3.4}:
The pizza is not good enough.

Quadruples:

pizza-Food\#General-not good enough-negative

Explanation: "not good enough" expresses the negation of the positive sentiment and needs to be annotated in the opinion.

\section{Category Definition}
\label{appendix:category_definition}

We have defined a new category for the Clothing domain, which includes the following entity labels: Clothing, Top, Bottom, Shoes, Socks, Service, and attribute labels: General, Brand, Quality, Size, Looking, Option, and Others. 

Among them, Quality mainly includes material properties such as washability, wear resistance, pilling resistance, smoothness, durability, warmth, lightness, comfort, etc.; Size describes whether the purchased product is big or small, the elasticity, the accuracy of the size chart and whether it fits well; Looking describes the product's color, design, etc.; Option represents the variety of styles and colors of the product.

\end{document}